\documentclass[letterpaper, 10pt, conference]{ieeeconf}
\IEEEoverridecommandlockouts                       
\usepackage{microtype}
\usepackage{graphicx}
\usepackage[font=small]{caption} 
\usepackage[font=small]{subcaption}
\usepackage{booktabs}
\usepackage[hidelinks]{hyperref}
\usepackage{listings}
\usepackage{float}
\usepackage{amsmath,amsfonts,amssymb}
\usepackage{optidef}

\usepackage{yaacro}
\usepackage{tabularx}
\usepackage{cite}
\usepackage{algpseudocode}
\usepackage{soul}
\usepackage[colorinlistoftodos]{todonotes}

\usepackage[ruled, vlined, linesnumbered]{algorithm2e}

\floatname{algorithm}{Procedure}

\newcommand\footnoteref[1]{\protected@xdef\@thefnmark{\ref{#1}}\@footnotemark}
\newcommand{\fixlabels}[1]{\textcolor{purple}{[FIX THE LABELS!]}}

\title{\LARGE \bf Intermittent Rendezvous Plans with Mixed Integer Linear Program for Large-Scale Multi-Robot Exploration}

\author{
    Alysson Ribeiro da Silva$^{1}$ and Luiz Chaimowicz$^{1}$
    \thanks{
    }
    \thanks{
        Alysson Ribeiro da Silva and Luiz Chaimowicz are with the VeRLab, Universidade Federal de Minas Gerais, Brazil. 
        {
            \tt\small \{alysson.silva,chaimo\}@dcc.ufmg.br
        }
    } 
}

\begin{document}

\maketitle
\thispagestyle{empty}
\pagestyle{empty}

%%%%%%%%%%%%%%%%%%%%%%%%%%%%%%%%%%%%%%%%%%%%%%%%%%%%%%%%%%%%%%%%%%%%%%%%%%%%%%%%
\begin{abstract}
Multi-Robot Exploration (MRE) systems with communication constraints have proven efficient in accomplishing a variety of tasks, including search-and-rescue, stealth, and military operations. While some works focus on opportunistic approaches for efficiency, others concentrate on pre-planned trajectories or scheduling for increased interpretability. However, scheduling usually requires knowledge of the environment beforehand, which prevents its deployment in several domains due to related uncertainties (e.g., underwater exploration). In our previous work, we proposed an intermittent communications framework for MRE under communication constraints that uses scheduled rendezvous events to mitigate such limitations. However, the system was unable to generate optimal plans and had no mechanisms to follow the plan considering realistic trajectories, which is not suited for real-world deployments. In this work, we further investigate the problem by formulating the Multi-Robot Exploration with Communication Constraints and Intermittent Connectivity (MRE-CCIC) problem. We propose a Mixed-Integer Linear Program (MILP) formulation to generate rendezvous plans and a policy to follow them based on the Rendezvous Tracking for Unknown Scenarios (RTUS) mechanism. The RTUS is a simple rule to allow robots to follow the assigned plan, considering unknown conditions. Finally, we evaluated our method in a large-scale environment configured in Gazebo simulations. The results suggest that our method can follow the plan promptly and accomplish the task efficiently. We provide an open-source implementation of both the MILP plan generator and the large-scale MRE-CCIC.
\end{abstract}

%%%%%%%%%%%%%%%%%%%%%%%%%%%%%%%%%%%%%%%%%%%%%%%%%%%%%%%%%%%%%%%%%%%%%%%%%%%%%%%%

\section{Introduction}

\begin{figure}[!t]
\includegraphics[width=1.0\linewidth]{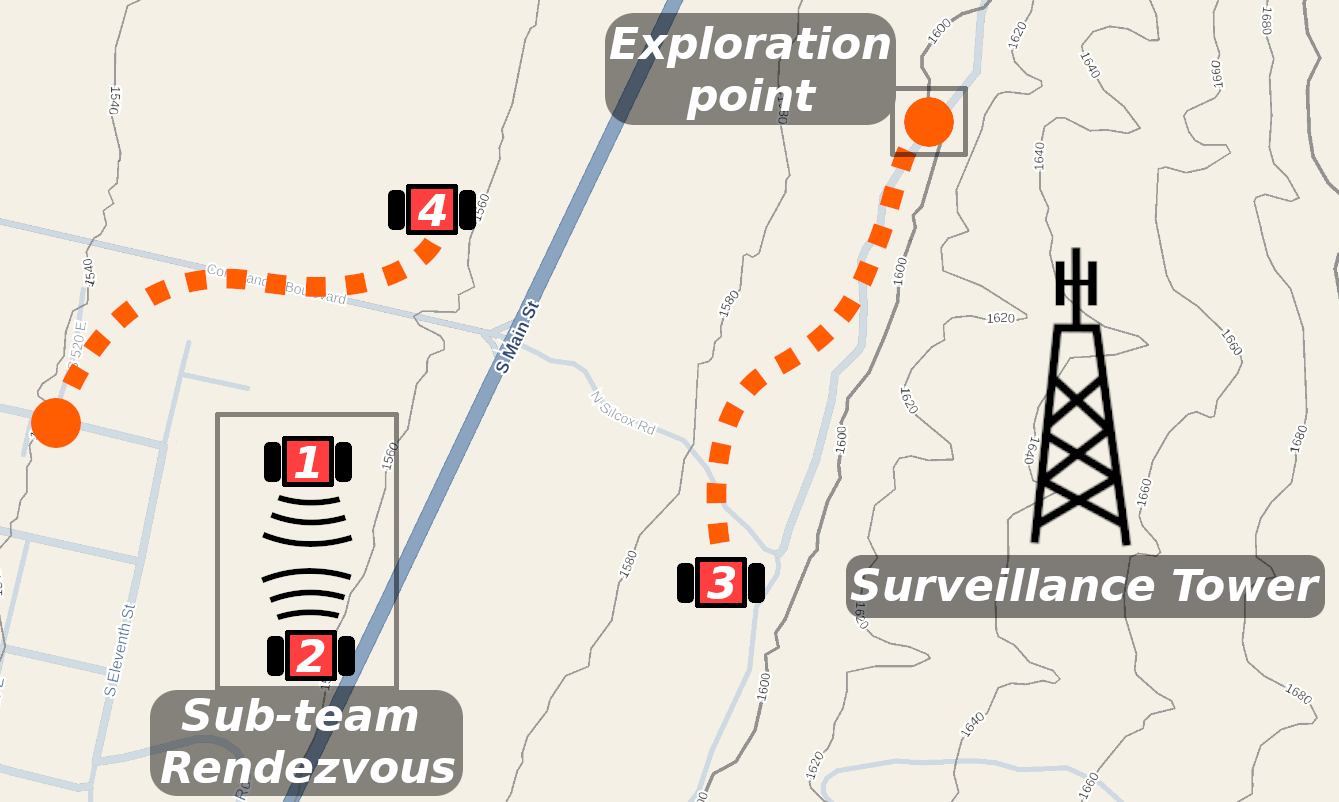}
\caption{Example of application in MRE with limited connectivity. Four robots performing an information gathering mission, trying to avoid being perceived by a \textit{Surveillance Tower}. To achieve this, they rely on scheduled rendezvous to exchange information using a low-range communications network. In this snapshot, robots 1 and 2 meet in a sub-team rendezvous while robots 3 and 4 explore the area through exploration points. The scheduled rendezvous must occur promptly to allow a base of operations to predict how the mission unfolds, which fosters human-robot collaboration.}
\label{fig:stealth}
\vspace{-2.0em}
\end{figure}

Researchers have been applying Multi-robot Exploration (MRE) with limited connectivity concepts for various tasks over the years. For example, some tasks require robots to be stealth \cite{Teng1993, Park2009}, allocate resources \cite{Goarin2024, Xu2025}, maintain connectivity \cite{Tang2024}, and explore unknown areas known as frontiers \cite{Bramblett2022} as we exemplify in Fig.~\ref{fig:stealth}. In contrast, others involve human-robot collaboration in unpredictable underwater currents \cite{Khatib2016, Khatib2022}. These scenarios can result in limited communication ranges \cite{Akyildiz2015}, posing several challenges for the coordination and information sharing of multi-robot systems. 

In the absence of global communication, robots must share information either at predetermined rendezvous points or through opportunistic encounters to accomplish connectivity requirements. For instance, \cite{Derek2024, Pongsirijinda2025} proposes opportunistic methods that allow robots to explore an environment while sharing information when they meet by chance, keeping a certain degree of intermittent communication. Differently, \cite{Bramblett2022} proposes a method where robots explore and decide to share information when appropriate. A different approach is to schedule a mission that is interpretable, such as the methods of \cite{Hollinger2012, Kantaros2019, Nore2025, Pongsirijinda2025}. In some of these methods, robots follow trajectories defined beforehand in a known environment, keeping intermittent communication. Most opportunistic strategies fail to deliver requirements such as interpretability, while the scheduling-based ones fail in MRE missions due to the lack of information regarding the environment. To address those issues for MRE missions, and yet achieve a certain degree of interpretability, in our previous work \cite{Silva2024}, robots execute an arbitrary exploration policy while following a rendezvous plan. This policy allows them to adapt where they should meet, considering unpredictable circumstances (e.g., spreading rendezvous locations dynamically), and makes them intermittently connected over time. However, it had two significant drawbacks: 1) it generates rendezvous plans through a meta-heuristic, which is non-optimal and harder to extend; 2) it was primarily designed to run in grid-world simulations, which were hard to translate to real deployments.  

% All these methods are opportunistic, due to their lack of prior planning, which adds significant limitations in terms of human-team collaboration since they are not interpretable with ease and portray unpredictable behavior. A different approach is to schedule a mission that is interpretable, such as the methods of \cite{Hollinger2012, Kantaros2019}. In these methods, robots follow trajectories defined beforehand in a known environment, keeping intermittent communication. However, they are not suited for exploration, because new information acquired during the mission alters the initial optimal planning. In this regard, our previous work \cite{Silva2024} deals with this issue for MRE missions by letting robots execute an arbitrary exploration policy while following a rendezvous plan. This policy delivers a certain degree of adaptability under unpredictable circumstances (e.g., spreading rendezvous locations dynamically, forcing exploration in specific regions) and makes robots intermittently connected over time. However, it had two major drawbacks: 1) the method generates rendezvous plans through a meta-heuristic, which is non-optimal and harder to extend; 2) the method was primarily designed to run in a grid-world simulation, which was hard to translate to real deployments.

In this work, we handle these drawbacks by proposing the Multi-Robot Exploration with Communication Constraints and Intermittent Connectivity (MRE-CCIC) problem, providing a linearized MILP formulation to generate optimal rendezvous plans and a policy to follow them. Secondly, we provide a Rendezvous Tracking in Unknown Scenarios (RTUS) mechanism that makes robots meet when assigned, considering an optimal local planner. Our method fosters human-robot collaboration in missions such as~\cite{Khatib2016} by leveraging the ability to predict when rendezvous encounters are going to happen, and conveying information sharing in the absence of larger communication ranges~\cite{Akyildiz2015}. Furthermore, the MRE-CCIC is also suited for applications where broadcasting robots' positions is not allowed by the mission's requirements (e.g., stealth operations). We evaluated the improvements in a large-scale ($65000$ $m^2$) ROS simulation with optimal local planners alongside a naive frontier exploration method~\cite{Bramblett2022}. The results suggest that our method enables robots to explore and meet when designated, considering realistic and efficient trajectories from the local planners in simulations with predictable behavior.

Our contributions are summarized as follows.

\begin{itemize}
    \item We provide a MILP (Mixed Integer Linear Program) to represent and generate optimal rendezvous plans in a multi-robot exploration setting, where robots must explore and share information at the designated locations.
    \item We propose a Rendezvous Tracking in Unknown Scenarios mechanism that allows robots to follow the generated plan, considering realistic trajectories on time.
    \item We provide an open-source implementation of the plan generator from the MILP formulation.
    \item We also provide the open-source ROS framework for large-scale exploration with intermittent connectivity.
\end{itemize}

\section{Related Works}

\subsection{Opportunistic Interactions VS Pre-planned Rendezvous}

We consider two forms of opportunistic interactions for intermittent communications in robot networks. In the first one, when two robots can establish a communication link by chance, they can deliberately meet at a specific location to exchange information, or do it while navigating. Differently, the second one implies that robots do not take advantage of a communication link established by chance, but rather continue their routines, trying to broadcast all they have to other connected robots. Most, if not all, methods we investigated fall into the second form. For instance, \cite{Derek2024} proposes a Deep Reinforcement Learning (DRL) decentralized exploration policy where robots explore unknown regions represented by a graph and can engage in rendezvous by chance.
In contrast, \cite{Pongsirijinda2025} proposes an exploration strategy for robots with global communication based on entropy fields. In both methods, robots meet by chance due to the uncertain nature of exploration tasks, which the authors call implicit rendezvous encounters. Among the advantages of opportunistic approaches, there is freedom, which implies that the burden of other tasks does not constrain robots' exploration capabilities. Consequently, they could belong to a family of algorithms that are concerned solely with multi-robot exploration, such as the one proposed by \cite{yamauchi1999}. Their performance seems tied to how robots select places to visit (e.g., market-based \cite{Zlot2002}, frontier-based \cite{Bramblett2022}, Deep Reinforcement Learning \cite{Tan2023}). In particular, they are also hard to interpret and portray a degraded capacity for requirements such as interpretability (i.e., the ability to understand what is going on) and timely mission tracking. Allowing a ground human team to track and interpret the mission precisely, understanding the role and responsibility of each robot promptly, and receiving specific pieces of information as planned are more realistic scenarios for real applications (e.g., inspection, search-and-rescue) that exemplify the qualities of interpretability.

On the other hand, pre-planned rendezvous are more controllable and allow robots to execute their tasks while sharing information, given a set of constraints related to the mission assignment. For example, real-world application constraints can relate to obeying a time budget, energy costs, recharging demands, delivering information to a ground team, deliberately acting as a data mule at specific moments, or performing other timely tasks while other robots explore. The requirements above influence deployment costs that a human team could mitigate before the mission starts. For instance, the authors in \cite{Hollinger2012} and \cite{Hannes2020} propose solutions based on scheduled encounters through joint-path optimizations, which makes robots meet and exchange information repeatedly during the mission. To achieve this, they rely on connectivity graphs \cite{Bullo2009}. Similarly, Kantaros et al.~\cite{Kantaros2019} propose an intermittent communication method that divides robots into sub-teams. Members in different sub-teams must meet at rendezvous locations periodically following a predetermined schedule. Although their methods rely on global communication and the environment is known beforehand, they provide estimates from the path and trajectory planning phases. Most importantly, and unlike opportunistic approaches, their method is interpretable with ease. 

Recently, \cite{Cao2021, Cao2023} proposed a framework for complex 3D environments, which was able to fully explore the Satsop Nuclear Plant. In particular, these contributions aim for complex multi-level environment exploration through global sub-space tours by visiting key points considering a local visibility. Compared to other approaches, their method eliminates the necessity of metric map planning and allows for exploring facilities with several floors. The authors also extended their initial environment representation and tour follow policy with a multi-robot exploration policy that aims to maximize the likelihood of passing through other robots' tours. While maximizing the likelihood of meeting other robots, this event happens by chance, which does not ensure the requirements of a communication link among robots and a ground team and fails for a mission wich such a link. The authors compare their method with a generalized multi-point rendezvous strategy, where robots meet at several different points. Their results show efficiency gains compared to the multi-point rendezvous, which corroborates the fact that opportunistic strategies can be beneficial in terms of exploration time while devoid of more realistic communication requirements for human-robot collaboration in realistic scenarios.

\subsection{Joint-path Planning Issues and Scheduling}

The main issue with joint-path planning methods is the fact that robots must know the environment beforehand, which makes them impractical for any MRE mission. A different approach for MRE missions is scheduling. A common and optimal way to schedule interpretable and precise actions in robotic systems is to solve modified versions of the Job-shop Scheduling Problem (JSSP), where robots are machines and jobs represent their assigned tasks (i.e., navigate through the environment, find victims, meet other robots, explore frontiers). The counterpart of this approach is that robots must follow the scheduled task plan on time without knowing if the other robots were able to accomplish their role due to connectivity constraints. As an example of scheduling, \cite{Nore2025} recently proposed a Mixed Integer Linear Program (MILP) to solve a variant of the Vehicle Routing Problem (VRP) whose scheduled solutions allow a convoy of robots to share information at designated locations while accomplishing tasks at key points in the assigned mission area. Particularly in this work, the robots know the entire mission area beforehand. In contrast, in our previous work \cite{Silva2024}, we investigate scheduling rendezvous events for unknown environments, which turns out to be a hard problem to solve due to all the navigation and exploration uncertainties that prevent robots from following a solution of a VRP variant.
\begin{figure*}[!th]
    \centering
    \includegraphics[width=0.9\linewidth]{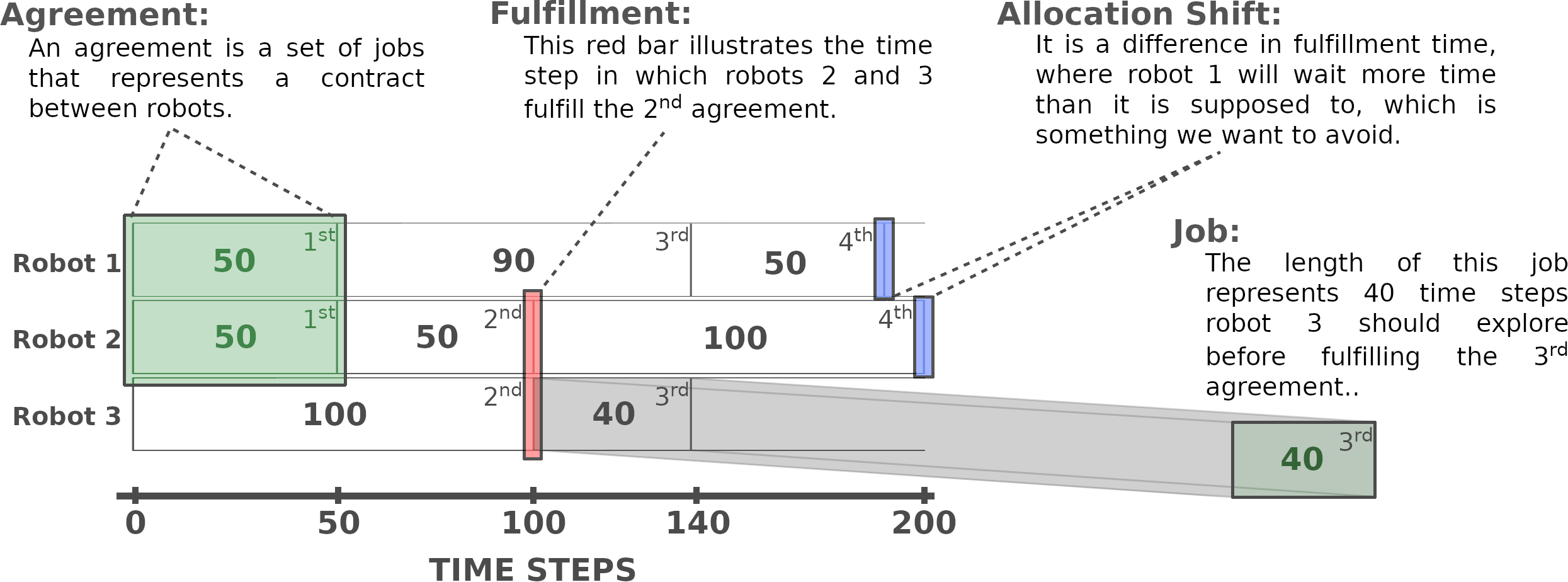}
    \caption{Scheduled rendezvous plan represented by coupled exploration jobs. In this example, robots $1$ and $2$ must meet after 50 time steps exploring the environment, because they are coupled by the jobs marked as the $1^{st}$ schedule. We call these coupled jobs agreements between robots, and they are analogous to humans exploring a place and establishing their meeting points based on the information they gather.}
    \label{fig:jssp_plan}
    \vspace{-2em}
\end{figure*}

\section{Problem Statement}

In this work, we propose the MRE-CCIC (Multi-robot Exploration with Communication Constraints and Intermittent Connectivity) problem as follows. Robots without global communication must acquire new information regarding their surroundings, detect points of interest to be visited, and share their findings with other robots, equipment, or a ground team, since their communication is limited. Robots have new points of interest to be visited unpredictably and must maintain intermittent connectivity and share their findings with other robots.

Robots can synchronize their tasks by solving variants of the VRP with synchronization points, ensuring that they can communicate at some points \cite{Soares2024}. However, the MRE-CCIC's inherent nature requires a Traveling Salesman Problem (TSP) to be solved every time new information is acquired. Furthermore, since they cannot communicate all the time, a VRP solution that considers global communication would fail as soon as they lose communication and obtain information not used during the planning phase.

Due to the limitations of VRP methods for exploration tasks, instead of depots and sequential tasks for synchronization points \cite{Nore2025}, we used our previous methodology based on coupled jobs and a multi-task policy in a DEC-POMDP \cite{Silva2024} to solve the MRE-CCIC.

Fig.~\ref{fig:jssp_plan} shows an instance of our proposed rendezvous scheduling mechanism that robots use to track their exploration tasks and sub-group rendezvous. In this example, each robot is a machine, and each job represents a task of exploring the environment (e.g., perceive, detect new points of interest, navigate, avoid collision). When a robot finishes executing a job (e.g., by achieving its maximum allowed exploration time or after acquiring enough information), it must meet with other robots to share its findings. To identify which robots must meet, we propose grouping jobs to represent a sub-team of robots. In our example, robot $1$ meets with robot $2$ after $50$ time steps of exploration specified by the $1^{st}$ schedule. Importantly, the set of all jobs and their processing times must obey the mission assignment constraints, such as maximum processing times and the available horizon.

\section{Modeling}

We define a solution to the MRE-CCIC as the interaction between $R$ robots in $S$ rendezvous, representing rendezvous meetings, during the mission assignment. Throughout the model, we use the subscript $i$ to represent the $ith$ rendezvous event and the subscript $j$ to represent the $jth$ robot. For the sake of simplicity in some explanations, we define a job processing time as $job_{i,j} = (a,b)$, where $a,b \in \mathbb{R}$, $a$ is its starting time, and $b$ is its ending time. If $a=0$ and $b=0$, then a job $job_{i,j}$ from a rendezvous $i$ does not exist. For any matrix $K$, we denote $k_i \in K$ as the entire row $i$ of $K$. We define matrix dimensions as superscripts, whereas subscripts define value ranges. We use standard set notation for constraining elements. For example, $K \in \mathbb{R}_+^{N\times N}$ defines a positive real-valued matrix with $N\times N$ elements. Finally, we use the operator $\cdot$ to denote scalar multiplication. Next, we present our linear MILP to generate job allocations for the MRE-CCIC.

Consider a rendezvous matrix as $K \in \{0,1\}^{S \times R}$ and the jobs, starting and ending times, matrices as $S \in \mathbb{R}_+^{S \times R}$ and $E \in \mathbb{R}_+^{S \times R}$, respectively. For example, $k_{i,j} \in \{0,1\}$ from $K$ is $1$ if robot $j$ participates in the rendezvous $i$ and $0$ otherwise. Analogously, the elements $s_{i,j} \in S$ and $e_{i,j} \in E$ are positive real values that represent the start and ending times of job $j$ at rendezvous $i$, respectively. We proposed this convention in our previous work because it was easier to identify and group robots that participate in the same rendezvous. We propose a latest available allocation matrix $L \in \mathbb{R}_+^{S \times R}$, where each $\ell_{i,j} \in L$ represents the allowed starting time of a $job_{i,j}$, an auxiliary array $H = \{h_1,...,h_S\}$, $H \in \mathbb{R}_+$, whose each element is a positive real value that represents the highest ending time of a rendezvous group. We define the number of allocations for a rendezvous $i$ as $a_{i} = \sum_{j}^R k_{i,j}$ and the number of jobs a robot $j$ participates as $p_{j} = \sum_{i}^S k_{i,j}$. 

\subsection{Rendezvous Locks and Latest Available Allocation}

Since it may be difficult to track the time when a robot is available for exploration, our model defines each element from $L$ with Eq.~\ref{eq:max_latest}, where $i-l$ represents the last rendezvous in which robot $j$ participates.

\begin{equation}
\label{eq:max_latest}
l_{i,j} =
  \begin{cases}
    \max(e_{i-l})       & \quad \text{if } k_{i-l,j} = 1 \\
    0  & \quad \text{otherwise}
  \end{cases}
\end{equation}

Analogously, to implement the logic of Eq.~\ref{eq:max_latest} as an integer program, we have the $H$ array, whose each element is a positive real value that represents the highest ending time of a rendezvous group. For example, in the following $E$ matrix instance

\vspace{-1.0em}
\begin{align*}
E = \begin{bmatrix}
        0 & 10 & 10 \\
        50 & 50 & 100 \\
    \end{bmatrix}
\end{align*}

\noindent
the related $H$ array is represented by $[10,100]$, where each element $h_k \in H$ is the highest ending time of the $kth$ rendezvous. For instance, the highest ending time of the first rendezvous is $h_1 = 10$ and the highest ending time of the last rendezvous is $h_2 = 100$.

\subsection{Objectives}

Our objective is to minimize the total work done error ($W_{err}$) and the deviation between processing times ($J_{err}$).
Since our problem is an elementary non-linear due to job allocations constrained on the decision variables, we linearized all objectives with the L1 norm, which is a proper metric and a common ground in linear programming.

We approximate the total work done error with Eq.~\ref{eq:total_workdone}.

\begin{equation}
\label{eq:total_workdone}
    W_{err} = \left(\sum_{i=1}^S\sum_{j=1}^R |e_{i,j} - s_{i,j}|\right) - m_{assign}
\end{equation}

\noindent
where, $m_{assign}$ is the mission assignment temporal budget.

Differently, our method is aware of the dispersion among processing times through the approximation defined by Eq.~\ref{eq:proc_dev}.

\begin{equation}
    \label{eq:proc_dev}
    J_{err} = \frac{1}{S\cdot R}\sum_{i=1}^S\sum_{j=1}^R (|e_{i,j} - s_{i,j}| - \bar{\tau})
\end{equation}

\noindent 
where, $\bar{\tau}$ is the average job length.

\subsection{Formulation} 
Next, we present our model, given our objectives, which are to minimize the total work done error, the idle time error, and the approximate processing times.

\begin{subequations}\label{eq:bounds}
\begin{align}
    \min \quad & \alpha W_{err} + \beta J_{err}\notag\\
    \text{s.t.} \notag\\
    %
    %%%%%%%%%%%%
    % highest allocation constraints
    h_i                 & \geq e_{i,j} - M \cdot (1 - k_{i,j})               \quad \forall i,j\tag{a1}\label{eq:highest_ending_linearization_1}\\
    %
    %%%%%%%%%%%%
    % latest allocation constraints
    \ell_{i,j}                   & \geq \ell_{i-1,j} - M \cdot k_{i-1,j}     \quad \forall i>1,j\tag{b1}\label{eq:mips_hack_to_propagate_values_1}\\
    \ell_{i,j}                   & \leq \ell_{i-1,j} + M \cdot k_{i-1,j}     \quad \forall i>1,j\tag{b2}\label{eq:mips_hack_to_propagate_values_2}\\
    \ell_{i,j}                   & \geq h_{i-1} - M \cdot (1-k_{i-1,j})      \quad \forall i>1,j\tag{b3}\label{eq:mips_hack_to_propagate_values_3}\\
    \ell_{i,j}                   & \leq h_{i-1} + M \cdot (1-k_{i-1,j})      \quad \forall i>1,j\tag{b4}\label{eq:mips_hack_to_propagate_values_4}\\
    s_{i,j}                      & \geq \ell_{i,j} - M \cdot (1-k_{i,j})     \quad \forall i>1,j\tag{b5}\label{eq:mips_hack_to_propagate_values_5}\\
    %
    %%%%%%%%%%%%
    % job selection constraints
    s_{i,j}           & \leq M \cdot k_{i,j}                               \quad \forall i,j\tag{d1}\label{eq:decision_linearization_1}\\
    e_{i,j}           & \leq M \cdot k_{i,j}                               \quad \forall i,j\tag{d2}\label{eq:decision_linearization_2}\\
    %%%%%%%%%%%%
    % job length contrainsts
    e_{i,j}           & \geq s_{i,j} \quad \forall i,j\tag{c1}\label{eq:proc_time_1}\\
    e_{i,j}           & \leq m_{assign}   \quad \forall i,j\tag{c2}\label{eq:proc_time_2} \\
    e_{i,j} - s_{i,j}          & \leq M \cdot k_{i,j}                               \quad \forall i,j\tag{c3}\label{eq:proc_time_3}\\
    e_{i,j} - s_{i,j}           & \geq min_{proc} \cdot k_{i,j}                      \quad \forall i,j\tag{c4}\label{eq:proc_time_4}\\
    %%%%%%%%%%%%
    % min number of jobs per robots, min robots er schedule, max robots per schedule
    p_{j}             & \geq 1                               \quad \forall j\tag{e1}\label{eq:allocation_bounds_1}\\
    a_{i}             & \geq min_{robots}                     \quad \forall i\tag{e2}\label{eq:allocation_bounds_2}\\
    a_{i}             & \leq max_{robots}                     \quad \forall i\tag{e3}\label{eq:allocation_bounds_3}\\
    a_{i}             & \leq R - 1                                \quad \forall i\tag{e4}\label{eq:allocation_bounds_4}\\
    %%%%%%%%%%%%
    % general bounds
    \ell_{i,j}, h_i & \geq 0                                    \quad \forall i,j\tag{f1}\label{eq:bounds_1}\\
    s_{i,j}, e_{i,j} & \geq 0                                    \quad \forall i,j\tag{f2}\label{eq:bounds_2}\\
    s_{1,j}, \ell_{1,j} & = 0                                       \quad \forall j\tag{f3}\label{eq:bounds_3}\\
    k_{i,j} & \in \{0,1\}                               \quad \forall i,j\tag{f4}\label{eq:bounds_4}
\end{align}
\end{subequations}

All objectives were linearized with the big-M method as done by \cite{Nore2025} and $\alpha$ and $\beta$ are used to try prioritizing work done or homogeneous job sizes, respectively. Constraint (\ref{eq:highest_ending_linearization_1}) ensures that the latest job ending for rendezvous $i$ is greater among all $j$ used to set new jobs starting times. Differently, constraints (\ref{eq:mips_hack_to_propagate_values_1}) to (\ref{eq:mips_hack_to_propagate_values_5}) implement a value propagation logic inside the matrix $L$, which helps to know the next available allocation for each robot separately. Next, constraints (\ref{eq:decision_linearization_1}) to (\ref{eq:decision_linearization_1}) ensures that non-selected jobs to have zero length. Constraints (\ref{eq:proc_time_1}) to (\ref{eq:proc_time_4}) ensure bounds for job lengths. In particullar (\ref{eq:proc_time_2}) prevents the mission to end outside the horizon $m_{assign}$. Constraints (\ref{eq:allocation_bounds_1}) to (\ref{eq:allocation_bounds_4}) implements allocation bounds defining the maximum and minimum allowed jobs per robot and robots per job. Finally, constraints (\ref{eq:bounds_1}) to (\ref{eq:bounds_4}) represent general boundaries.

\subsection{Intuition Behind the Latest Allocation Matrix}

In particular, we abuse the big-M to keep track of the next available job starting time in the $L$ matrix. The intuition of the related constraints is to provide $l_{i,j}$ with two choices of values. On the first choice, it receives the last allocated value for its robot or $l_{i,j} = l_{i-1,j}$, which ensures value propagation. Otherwise, if robot $j$ gets allocated to rendezvous $i-1$, then $l_{i,j} = h_{i-1}$, which ensures that it does not need to propagate any value from the previous allocation.

\subsection{Rendezvous Following and Exploration Policies}

In MRE-CCIC, robots must have the freedom to explore the environment while keeping track of the planned rendezvous.
The exploration policy is composed of two parts: 1) rendezvous following policy; 2) naive frontier exploration.
The general idea is that all robots must keep track of a global clock, which is easy to accomplish in a distributed manner by setting a standard reference on each robot before the mission starts, while executing the jobs from the rendezvous plan (i.e., exploring a frontier, navigating to a rendezvous location, and waiting for other robots).

Each robot keeps track of the rendezvous plan and which job they are supposed to execute now.
Each robot then explores the environment with a naive frontier exploration for the duration of the current job. To do so, robots gather data, create a metric representation, detect points of interest near unexplored areas (i.e., frontiers), and navigate to the one that maximizes their utility \cite{Bramblett2022}.
If a robot finds other robots by chance, it shares its maps, stops, and computes new points of interest with its newly acquired information. Different from our previous iteration \cite{Silva2024}, in this work, robots self-allocate to an appropriate place with a priority-based mechanism.
In this mechanism, each robot has a unique integer that helps identify it. When robots are in communication range, they have the same frontiers. Each robot sorts its frontiers by utility and picks the one associated with its identifier.
If a robot is exploring and the current mission time demands that it finish the current job and go to the assigned rendezvous location, it stops exploring and navigates to the designated zone. 
Once in the zone, it exchanges information, updates the rendezvous locations of the current sub-team with a method similar to \cite{Silva2024}, and starts executing the next job from the plan.
The stop conditions for this process are: 1) there are no more jobs to execute; 2) the total exploration time exceeds the mission's maximum allowed time budget; 3) there are no more places of interest to explore.

\subsection{Mitigating Rendezvous Waiting Times}

A significant issue with our previous work \cite{Silva2024} is the fact that we evaluated the method in a grid-world simulator with discrete time-steps. Differently, in this work, robots track the rendezvous plan and work with realistic trajectories with an optimal local planner based on elastic bands. In our model, jobs represent the action of exploring the environment, navigating, and meeting with other robots, because in MRE tasks, we often can not predict how the mission will unfold beforehand, making it hard to create a separate job for each sub-task. In this regard, we propose a \textit{Rendezvous Tracking in Unknown Scenarios} (RTUS) shown in Algorithm~\ref{alg:mreccic}. In this mechanism, robots keep track of the amount of time they have left before finishing the current job

\begin{equation}
    \Delta T_{window} = T_{rendezvous} - T_{current}
\end{equation}

\noindent
where $T_{rendezvous}$ is the mission time it must arrive at the current rendezvous location, and $T_{current}$ is the current mission time.
For each time step $t$, each robot computes the heuristic value

\begin{equation}
    \mathcal{H} = \Delta T_{window} - T_{path}
\end{equation}

\noindent
where $T_{path} = \frac{length}{\dot x}$ is the time it takes to reach the current rendezvous location $l$, considering an expected velocity $\dot{x}$ measured in meters per second. If $\mathcal{H} \leq 0$, then the robot must start navigating towards its next rendezvous location, which consequently minimizes the waiting times of all jobs at the assigned locations. This method implicitly represents the sub-task of navigating towards rendezvous locations in the current job. It is important to note that, with this method, robots can not navigate through unknown zones as proposed by \cite{Bramblett2022}, otherwise the calculations of $\mathcal{H}$ will result in unpredictable behavior.

\begin{algorithm}[t!]
\footnotesize
\DontPrintSemicolon
\KwIn{Plan $\mathcal{K}$, $S$, $E$}
\SetKwBlock{Begin}{function}{end function}
\Begin($\text{MRE-CCIC} {(\text{mission constraints }c)}$) {
    Receives $\mathcal{K}$, $S$, and $E$, given the mission constraints;\;
    Extract the rendezvous plan $P$ through \cite{Silva2024};\;
    \textbf{Should rendezvous} $\leftarrow$ false;\;
    \ForEach{time step $t$} {
       get current rendezvous location $l$ from $P$;\;
       \If{\textbf{Should rendezvous} given $t$} {
            Go to $l$ and wait until all other robots assigned to this encounter are there;\;
            \If{All robots assigned for this rendezvous are at $l$} {
                Assign new dynamic rendezvous locations through \cite{Silva2024} and obtain the next $l$;\;
                Continue with exploration;\;
                \textbf{Should rendezvous} $\leftarrow$ false;\;
            }
       } \Else {
            Explore frontiers;\;
            If during exploration, meet other robots by chance, compute new frontier set $F$;\;
            Sort $F$ by utility $U = \frac{value}{cost}$;\;
            Select $f_{id}$ to explore, where $id$ is this robot unique identifier;\;
       }
       Compute $T_{path}(l,r_{location})$ with the A* algorithm;\;
       Compute $\Delta T_{window} = T_{rendezvous} - T_{current}$;\;
       Compute $\mathcal{H} = \Delta T_{window} - T_{path}(l,r_{location})$;\;
       \If{$\mathcal{H} \leq 0$} {
            \textbf{Should rendezvous} $\leftarrow$ true;\;
       }
    }
    Go back to the initial location;\;
}
\caption{\small{Individual Robot MRE-CCIC with Rendezvous Mission Tracking}}
\label{alg:mreccic}
\end{algorithm}

\section{Experimental Configuration}

We evaluated our method MRC-CCIC against our previous work \cite{Silva2024}, which we refer to as Silva2024, since we are proposing a direct improvement to handle realistic trajectories when exploring large-scale areas. We evaluated the method through the following metrics: 1) Rendezvous Accomplishment Times; 2) Average Waiting Times; 3) Explored Area. The Rendezvous Accomplishment Times, measured in minutes, represent the time step in which robots reach the designated rendezvous locations during the assigned mission. With these experiments, we verify whether robots can explore the environment and reach the assigned rendezvous locations at the designated times, a capability that Silva2024 lacks. Differently, the Average Waiting Times, measured in seconds, help check how much time robots keep waiting at each rendezvous location. We use this test to verify how much time is lost waiting for other robots before proceeding with exploration. This metric also gives us insights into the method's capability to explore a certain amount of area given the maximum allowed mission time $m_{assign}$. Finally, the Average Explored Area metric, measured in squared meters as the average among all robots, shows whether the method can explore a certain amount of area even when arriving at the rendezvous locations at the designated times with the RTUS (Rendezvous Tracking in Unknown Scenarios). This metric is computed as an average because robots do not have global communication, which can highlight how quickly information spreads among robots.

\subsection{Simulation Environment}

\begin{figure}[!t]
    \centering
    \includegraphics[width=0.8\linewidth]{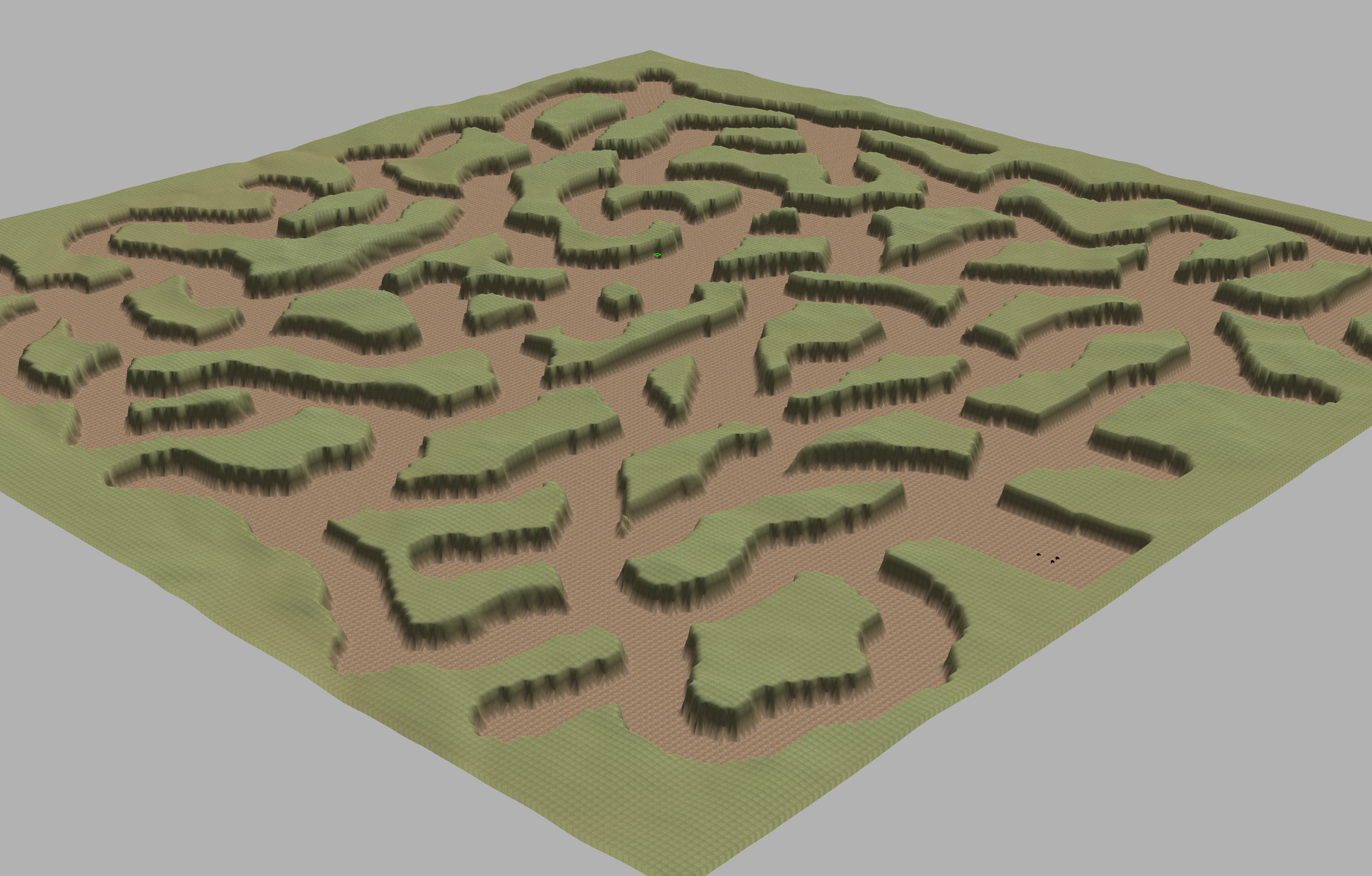}
    \caption{Large-scale environment we used to evaluate the performance of the RTUS and the optimal local planning.}
    \label{fig:mission_area}
    \vspace{-2.0em}
\end{figure}

\begin{figure}[!t]
    \centering
    \includegraphics[width=0.9\linewidth]{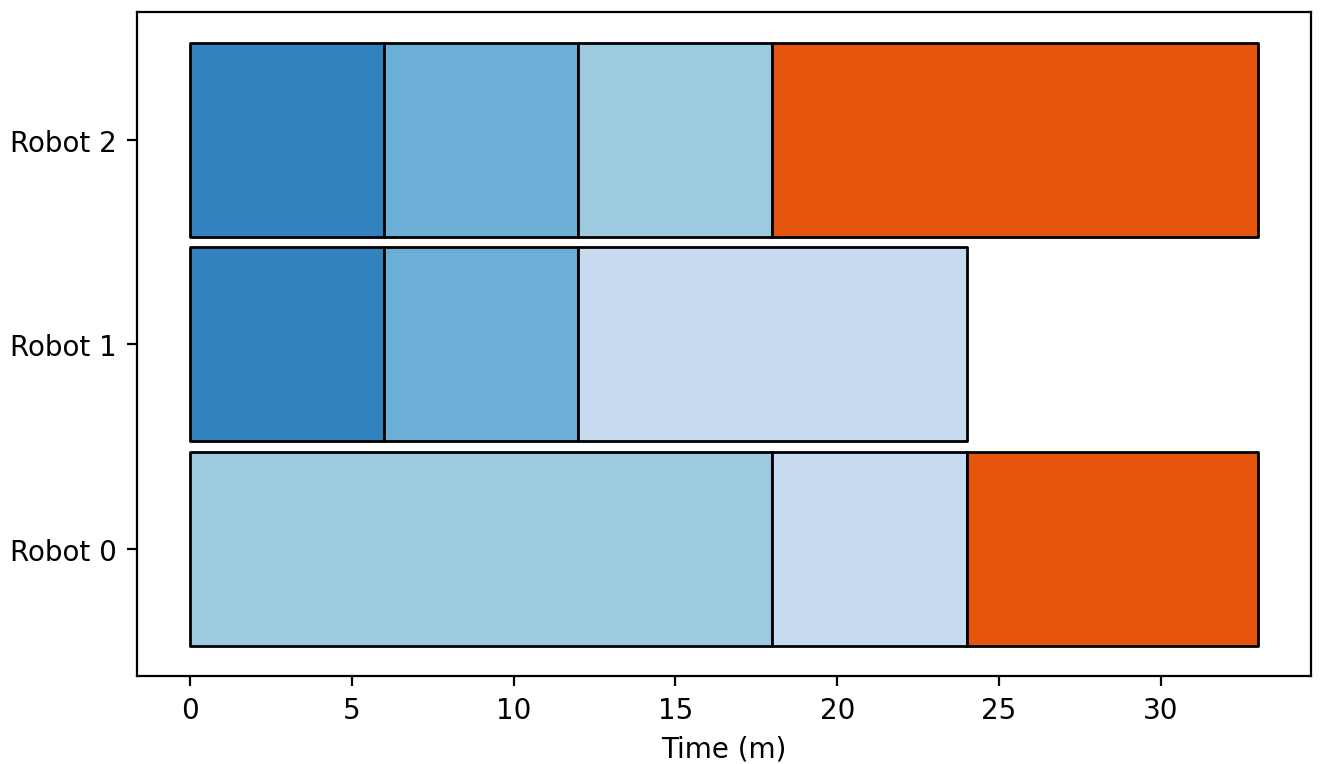}
    \caption{Rendezvous plan we generated for evaluation. It is comprised of $3$ robots and $5$ rendezvous encounters for an exploration mission of $30^{+5}_{-5}$ minutes. Each color represents a sub-team of robots. For example, robots $0$ and $1$ at minute $24$.}
    \label{fig:rendezvous}
    \vspace{-2.0em}
\end{figure}

In this work, we used the Robot Operating System (ROS) Gazebo for simulations. We set the MRE-CCIC mission assignment in the area we illustrate in Fig.~\ref{fig:mission_area} of roughly $65000$ $m^2$. For both methods, we generate a plan comprised of $5$ rendezvous and a maximum mission duration $30^{+5}_{-5}$ minutes from Fig.~\ref{fig:rendezvous}. We deployed $3$ Pioneer robots equipped with a 360-degree 2D LiDAR with a range of $100$ meters for localization and mapping, and four RGB cameras with a resolution of $192 \times 128$ for short-range obstacle avoidance, as it simulates deployments based on commertial solutions\footnote{\href{https://ouster.com/}{VLP16 LiDAR from ouster is a common solution which provide 100 m of coverage}}. The robots have a communication range of $10$ meters and use a communication model based on proximity. Most sensors we used employ gaussian noise with a standard deviation of $\pm 0.1$.

We reduced the resolution of the global map for better performance and performed map stitching to merge them when robots exchange maps. Our map stitching algorithm approach implements standard practices that consider the origins of Occupancy Grids and create a new one with an appropriate size. For mapping, we used the Octomap package and provided robots with perfect odometry, since this approach is more suited for efficient simulations. A SLAM or localization system can replace the odometry source without much effort, such as the SLAM Toolbox, FAST-LIO, gMapping, Global Navigation Satellite System (GNSS) based localization, and Extended Kalman Filter (EKF), and it is not detrimental for our evaluations. For navigation, we used move-base with an optimal local planner we configured to deliver a maximum traveling speed of $1$ m/s. We do not employ any traffic avoidance mechanism, since in a larger area, collisions are not common. To detect frontiers, we use a flood-fill algorithm. Robots compute their utilities as in \cite{Silva2024}. They always stop navigating after reaching a frontier to calculate new ones and avoid inconsistencies during the execution of the policy, which also helps spare processing power. To dynamically update the rendezvous locations and for decision making, we use a simple consensus rule based on the highest ID. For the metrics, we averaged the results between $5$ different runs, where robots start at a fixed location defined a priori at the map boundaries, and calculated their standard deviations. The source code for the project is available at this \href{https://github.com/multirobotplayground/Noetic-Multi-Robot-Sandbox}{github.com/multirobotplayground/Noetic-Multi-Robot-Sandbox}\footnote{\href{https://github.com/multirobotplayground/Noetic-Multi-Robot-Sandbox}{https://github.com/multirobotplayground/Noetic-Multi-Robot-Sandbox}}.

\section{Results}

\subsection{Estimate Velocity for Rendezvous Tracking in Unknown Scenarios}

\begin{figure}[!t]
    \centering
    \includegraphics[width=0.8\linewidth]{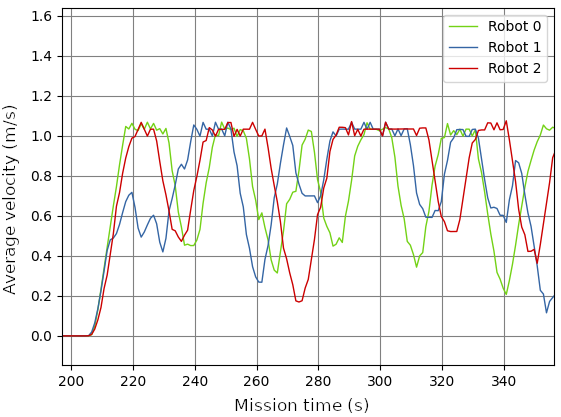}
    \caption{Velocity profile of $3$ robots when executing the mission to help decide the expected velocity of the RTUS (Rendezvous Tracking in Unknown Scenarios). The Y axis represents the velocities in m/s. The X axis shows the mission time in seconds.}
    \label{fig:velocities}
    \vspace{-1.5em}
\end{figure}

To decide which velocity to use when computing $\mathcal{H}$ from the RTUS, we collected the average velocity of all robots for the first minute of a mock mission using the Silva2024 method shown in Fig.~\ref{fig:velocities}. Each local minimum is associated with situations where robots stop navigating to compute new frontiers to spare processing power or rendezvous events. Differently, local maxima show when robots are exploring or navigating towards the assigned rendezvous. Given the observed profile, we defined $\dot x = 1$ m/s.

\subsection{Rendezvous Accomplishment Times}

\begin{figure}[!t]
    \centering
    \includegraphics[width=1.0\linewidth]{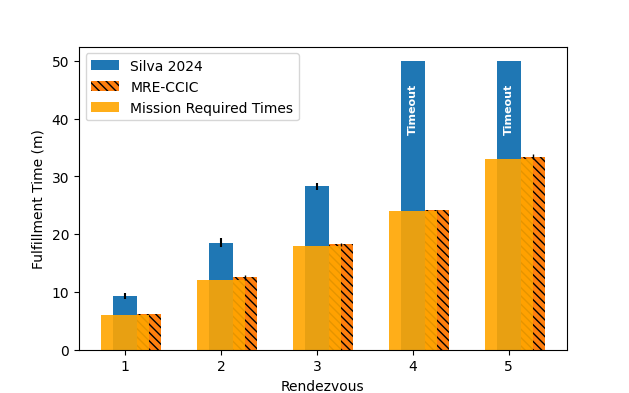}
    \caption{The rendezvous accomplishment times represent the time that robots arrived at the designated rendezvous location, given the plan we generated. The Y axis represents the mission time in minutes, while the X axis shows the number of rendezvous events in the order they occurred.}
    \label{fig:mission_accomplishment}
    \vspace{-2.0em}
\end{figure}

Fig.~\ref{fig:mission_accomplishment} shows the time at which robots meet at the assigned rendezvous locations. It is important to note that during the mission, rendezvous locations are updated according to the area already explored. The results suggest that our improvements allowed the robots to meet at the assigned locations in time, accomplishing the mission with an associated error within a few seconds. Differently, robots using Silva2024 were unable to meet at rendezvous $4$ and $5$, highlighted with a timeout label, and did not follow the plan on time. These results show that the MRE-CCIC was able to generate the mission plan, allows robots to explore as necessary for MRE tasks, and ensures they meet the mission assignment's times. As a consequence, our proposal can help a human-robot team to predict when each rendezvous event is going to happen, even when all the navigation and exploration uncertainties are under consideration, for scenarios where nothing abnormal happens (e.g., controller failures during the mission), respecting the mission's maximum allowed budget.

\subsection{Average Waiting Times}

\begin{figure}[!t]
    \centering
    \includegraphics[width=0.9\linewidth]{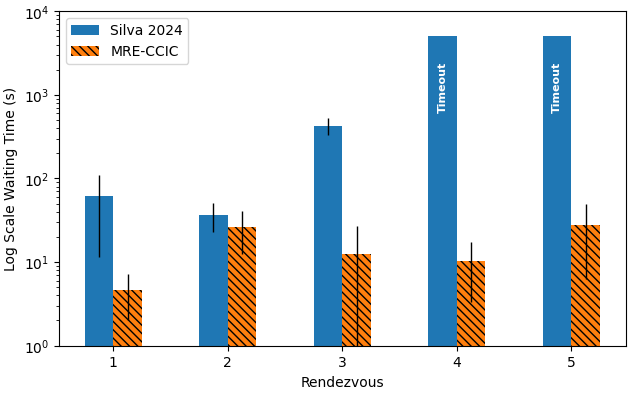}
    \caption{Waiting times at rendezvous locations represent how much time a sub-group of robots assigned to each rendezvous waits before the rendezvous is completed. The Y axis represents the wait time in seconds, and it is expressed on a log scale. The X axis represents the rendezvous event in the order in which they occurred.}
    \label{fig:waiting_times}
    \vspace{-1.0em}
\end{figure}

In Fig.~\ref{fig:waiting_times}, we show the average waiting time on a log scale. The results suggest that with an appropriate plan and simple mechanisms to execute it, robots can drastically reduce their waiting times at rendezvous locations. For instance, the MRE-CCIC won in all rendezvous events with waiting times in the scale of a few seconds. Differently, robots using Silva2024 took more than one minute before all assigned robots reached the assigned rendezvous. This indicates that MRE-CCIC can potentially achieve good performance in terms of explored area, even though robots spend part of their jobs navigating towards rendezvous locations, given the mission budget, because they wait less time at rendezvous locations.

\subsection{Explored Area}

\begin{figure}[!t]
    \centering
    \includegraphics[width=0.9\linewidth]{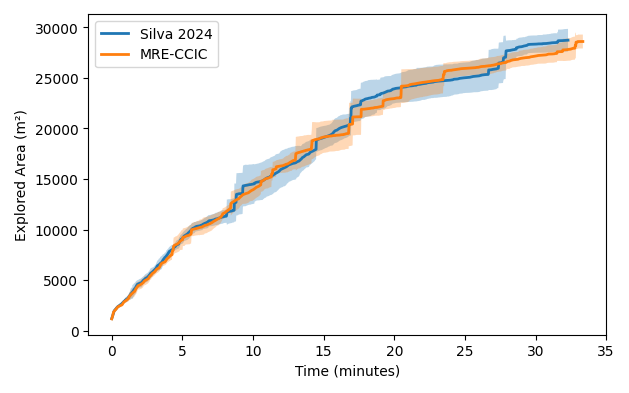}
    \caption{Explored area per minute represents the amount of area explored through the mission. The Y axis represents the area in $m^2$. The X axis represents the mission time in minutes.}
    \label{fig:area_explored}
    \vspace{-2.0em}
\end{figure}

\begin{figure}[!t]
    \centering
    \includegraphics[width=0.7\linewidth]{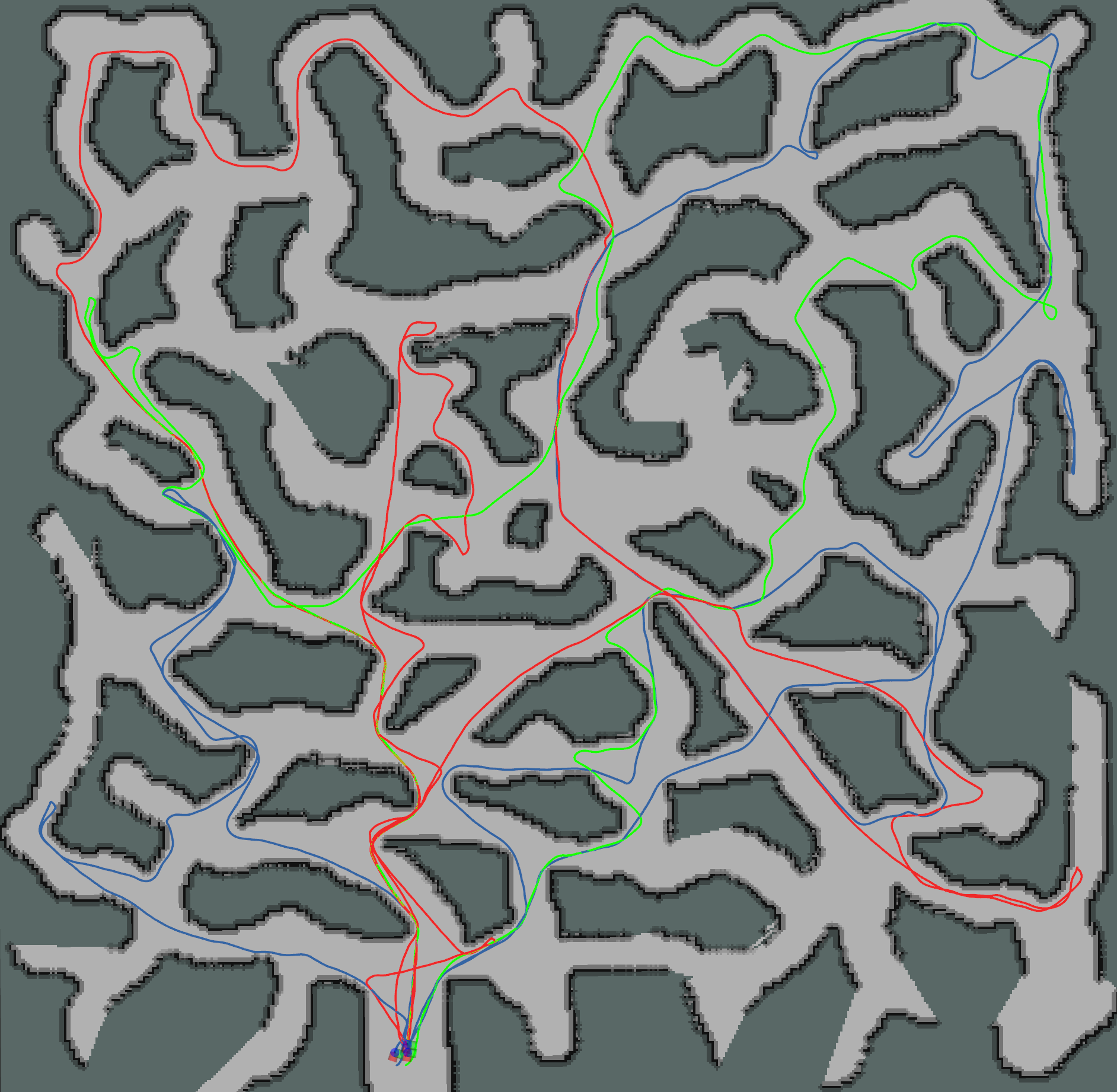}
    \caption{Trajectories and stitched maps at the end of the assigned mission from RViz from one run.}
    \label{fig:trajectories}
    \vspace{-2.0em}
\end{figure}

We show the explored area per time step in Fig.~\ref{fig:area_explored}. Our results suggest that the MRE-CCIC maintains the same performance as the Silva2024. During the mission assignment of $30$ minutes, both were able to explore roughly $27000$ $m^2$, which is very close to the area robots can reach in the environment. The trajectories of one run are shown in Fig.~\ref{fig:trajectories}. We could observe that the optimal local planning helps in straight movements and sharp corners, which increases the area explored per unit of time. We did not expect to see gains in performance from the MRE-CCIC and Silva2024, because both are using the same frontier exploration method in this evaluation. However, we were expecting the MRE-CCIC to perform worse, because it explores less due to the rendezvous tracking, which did not happen.

\section{Discussion and Conclusion}

In this work, we present an MILP (Mixed-Integer Linear Programming) formulation, which we refer to as MRE-CCIC (Multi-Robot Exploration with Communication Constraints and Intermittent Connectivity), for the exploration task. We also propose an RTUS (Rendezvous Tracking in Unknown Scenarios) to help robots accomplish their assigned rendezvous promptly, considering all the navigation and exploration uncertainties inherent in MRE tasks. It is essential to highlight that the MRE-CCIC was able to execute all rendezvous events predictably, and robots spent less time at the rendezvous locations. 

Among the limitations of the system, it is known that MIP formulation does not scale well with the number of robots, thus turning infeasible for larger teams. Our method does not addresses problems related to energetic constraints, which is crutial to real reployments. Furthermore, despite having full sensor noise simulations, we employed perfect odometry, which can lead to problems when integrating into real robots. Finally, our method provides a rendezvous allocation strategy for unknown scenarios and a policy integrated with an optimal local planner for dynamic obstacle avoidance instead of optimal visitation order, which is impossible to achieve for unknown scenarios in exploration tasks such as we defined in the MRE-CCIC problem.

Potential applications include mission assignments where robots must keep intermittent connectivity with other robots or a base station, fostering human-robot collaboration for scenarios such as oceanic exploration and stealth operations. The mission design can also respect time budgets (e.g., related to the time victims have available oxygen in a post-disaster site). It can provide the human team with estimates regarding when robots are going to meet and how much information has been acquired at specific moments in time, given prior knowledge about the environment. 

\section{Acknowledgment}

This work was partially funded by CAPES, CNPq, and FAPEMIG. We also acknowledge the use of GitHub Copilot as a programming assistant, primarily used as a search engine and to find bugs in code.

%%%%%%%%%%%%%%%%%%%%%%%%%%%%%%%%%%%%%%%%%%%%%%%%%%%%%%%%%%%%%%%%%%%%%%%%%%%%%%%%

\bibliographystyle{IEEEtran}
\bibliography{IEEEabrv, bib/exploration, bib/hazards, bib/rescue, bib/communication, bib/opt, bib/connectivity, bib/schedule_jssp, bib/benchmarks, bib/applications}

\end{document}